%% file: root.tex
\documentclass[letterpaper, 10 pt, conference]{ieeeconf}  

\IEEEoverridecommandlockouts                              

\usepackage{times}

\usepackage{multicol}
\usepackage[bookmarks=true]{hyperref}
\usepackage{graphicx}
\usepackage{svg}
\usepackage{amsfonts}
\usepackage{amsmath}
\usepackage{xcolor}
\usepackage{multirow}   
\usepackage{booktabs}
\usepackage[ruled,linesnumbered]{algorithm2e}

\title{\LARGE \bf
Learning Sampling Dictionaries for Efficient and Generalizable Robot Motion Planning with Transformers 
}

\author{Jacob J. Johnson$^{\dagger}$, Ahmed H. Qureshi$^{\ddagger}$, and Michael C. Yip$^{\dagger}$ 
\thanks{$^{\dagger}$J.J. Johnson and M.C.Yip are with the Electrical and Computer Engineering Department at University of California San Diego, La Jolla, CA, USA
{\tt\small \{jjj025, yip\}@eng.ucsd.edu}}
\thanks{$^{\ddagger}$ A.H. Qureshi is with the Department of Computer Science at Purdue University, West LaFayette, IN,  USA
{\tt\small ahqureshi@purdue.edu}}
}

\DeclareMathOperator*{\argmin}{argmin}

\begin{document}

\maketitle

\begin{abstract}
\input{1.Abstract}
\end{abstract}

\section{INTRODUCTION}
\input{2.Introduction}

\input{3.RelatedWorks}

\section{Background}
\input{4.Background}

\section{VECTOR QUANTIZED-MOTION PLANNING TRANSFORMERS}
\input{5.vq_mpt}

\section{EXPERIMENTS}
\input{6.Method}


\section{CONCLUSION}
\input{8.Conclusion}



\bibliographystyle{IEEEtran}
\bibliography{references}

\end{document}

%% file: 1.Abstract.tex
Motion planning is integral to robotics applications such as autonomous driving, surgical robots, and industrial manipulators. Existing planning methods lack scalability to higher-dimensional spaces, while recent learning-based planners have shown promise in accelerating sampling-based motion planners (SMP) but lack generalizability to out-of-distribution environments. To address this, we present a novel approach, Vector Quantized-Motion Planning Transformers (VQ-MPT) that overcomes the key generalization and scaling drawbacks of previous learning-based methods.
VQ-MPT consists of two stages. Stage 1 is a Vector Quantized-Variational AutoEncoder model that learns to represent the planning space using a finite number of sampling distributions, and stage 2 is an Auto-Regressive model that constructs a sampling region for SMPs by selecting from the learned sampling distribution sets.
By splitting large planning spaces into discrete sets and selectively choosing the sampling regions, our planner pairs well with out-of-the-box SMPs, generating near-optimal paths faster than without VQ-MPT's aid. It is generalizable in that it can be applied to systems of varying complexities, from 2D planar to 14D bi-manual robots with diverse environment representations, including costmaps and point clouds. Trained VQ-MPT models generalize to environments unseen during training and achieve higher success rates than previous methods. 
Videos and code are available at \url{https://sites.google.com/ucsd.edu/vq-mpt/home}.

%% file: 2.Introduction.tex
Sampling-based motion planning use randomly sampled points to generate a tree-based collision-free path between a start and goal locations \cite{doi:10.1177/02783640122067453, 508439}. However, random sampling is inefficient \cite{1242285} for goal-directed tasks, particularly when the search space spans a high number of dimensions. 
Since sampling-based motion planners (SMPs) are a fundamental component of numerous autonomous systems \cite{9561673, 1570348}, improving the efficiency and generalizability of the underlying planners enables these systems to handle more complex tasks that involve intricate sequences of planning, improves task execution, and reduces the need to retrain planners for different environments. While SMPs effectively generate a trajectory, they face several challenges in improving sampling efficiency. As the dimensionality of the configuration space increases, the "curse of dimensionality" makes sampling more difficult and time-consuming. Efficiently exploring high-dimensional spaces to find feasible paths is a significant challenge. These planners must also be able to reliably solve for different environments without the need for reconfiguring planner parameters. Most of these planners are probabilistically complete, i.e., the planner will find a path if a trajectory exists, given enough time. But finding a trajectory that is optimal, like the shortest path, is also a challenge. Numerous works have been proposed that address some of these challenges.

\begin{figure}
    \centering
    \includegraphics[width=\linewidth]{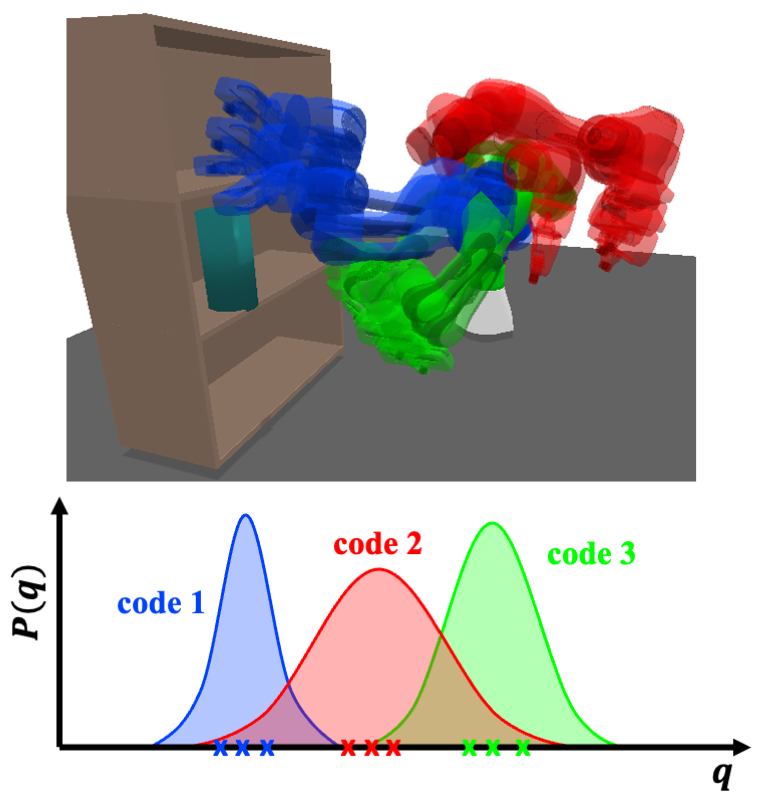}
    \caption{VQ-MPT can efficiently split high-dimensional planning spaces into discrete sets of distributions. Each distribution is represented using a latent variable called code or dictionary value. Given a planning problem, the model selects a subset of codes and samples from the associated distributions to construct the trajectory. By sampling efficiently, VQ-MPT reduces planning times by 2-6$\times$ compared to previous planners.}
    \label{fig:sample_fig}
    \vspace{-2em}
\end{figure}


%% file: 3.RelatedWorks.tex
For efficient sampling, prior works have reduced the search spaces through hand-crafted heuristics or parametric functions, decreasing planning time. The current state-of-the-art motion planners leverage goal-directed heuristics; Informed-RRT$^*$ (IRRT$^*$) \cite{6942976} and Batch Informed Trees (BIT$^*$) \cite{gammell2015batch} search for a path in an ellipsoidal region between the start and goal location. In \cite{qureshi2016potential, tahir2018potentially}, Artificial Potential Fields (APF) guide random samples toward regions with an optimal solution. Sampling-based A$^*$ \cite{doi:10.1177/02783640122067453} extends the A$^*$ search algorithm to sampling-based planning and uses heuristics to sample from selected vertices. But for higher dimensional spaces, sampling with these heuristics still leaves many samples unused for constructing a trajectory.

On the other hand, learning-based methods leverage data from prior planned data to accelerate planning in similar environments \cite{8412538, 9561104, 8653875, kumar2019lego}. Motion Planning Networks (MPNet) \cite{qureshi2019motion} was the first neural planner to generate the full motion planning solution through a recurrent sampling of its networks, given the current and goal position of the robot as well as the environment representation. MPNet considerably reduces planning time for higher dimensions, but these models do not generalize to larger environment representations \cite{mpt}. Other neural planners \cite{DBLP:conf/iclr/ChenDLYLS20, yu2021reducing} have also explored using neural networks for planning.

Transformer models are an ideal candidate for solving the planning problem because of their ability to make long-horizon connections \cite{NIPS2017_3f5ee243}. 
Advances in large language models, such as BERT \cite{DBLP:conf/naacl/DevlinCLT19}, and GPT \cite{NEURIPS2020_1457c0d6_GPT3}, have inspired similar efforts in solving planning tasks using transformer models \cite{chen2021decision_TransformerRL, janner2021sequence}. These models make better control decisions in robotic quadrupedal walking tasks by attending to proprioceptive and visual sensor data \cite{yang2022learning}. Although these works support the possibility of using transformer models for decision-making, it is difficult to interpret the policy's future control actions and provide any form of guarantee for the underlying planner. Other works \cite{mpt, chaplot2020differentiable} only solve for planar manipulators and 2D mobile robots because, inherently, their network models follow those used in image understanding in 2D discrete spaces. Since these models have to discretize the entire planning space, extending these methods to higher dimensional, continuous planning spaces would exponentially increase training and memory costs. Furthermore, these planners require the space in which the path is constructed (planning space) to overlap with the space in which the environment is represented (task space). For example, for a 14-degree-of-freedom bi-manual robot arm setup, the environment is represented using point clouds which is $\mathbb{R}^3$, while the planning space is $\mathbb{R}^{14}$. How these methods apply to environments with disjoint planning and task space is unclear.

In this work, we propose VQ-MPT, a scalable transformer-based model that accelerates SMP by narrowing the sampling space. VQ-MPT uses a Vector Quantized (VQ) model to discretize the planning space. VQ models are generative models with an encoder-decoder architecture similar to Variational AutoEncoder (VAE) models but with the latent dimension represented as a collection of learnable vectors referred to as dictionaries. A transformer model selects a subset of these learned vectors to generate the search region for the given planning problem. We describe in this paper how the VQ approach can be used in the context of motion planning, leading to the following major advantages:
\begin{enumerate}
    \item Reduces planning times by 2-6$\times$ compared to traditional planning algorithms such as BIT$^*$ and by 3-6$\times$ compared to learned planners such as MPNet.
    \item Scales to 14-dimensional planning spaces without compromising planning performance.
    \item Learns efficient quantization of high dimensional planning space without increasing the dictionary size.
    \item Generalizes to unseen in-distribution and out-of-distribution environments more successfully than learned planners such as MPNet.
\end{enumerate} 

%% file: 4.Background.tex
\begin{figure*}[t]
    \centering
    \includegraphics[width=\linewidth]{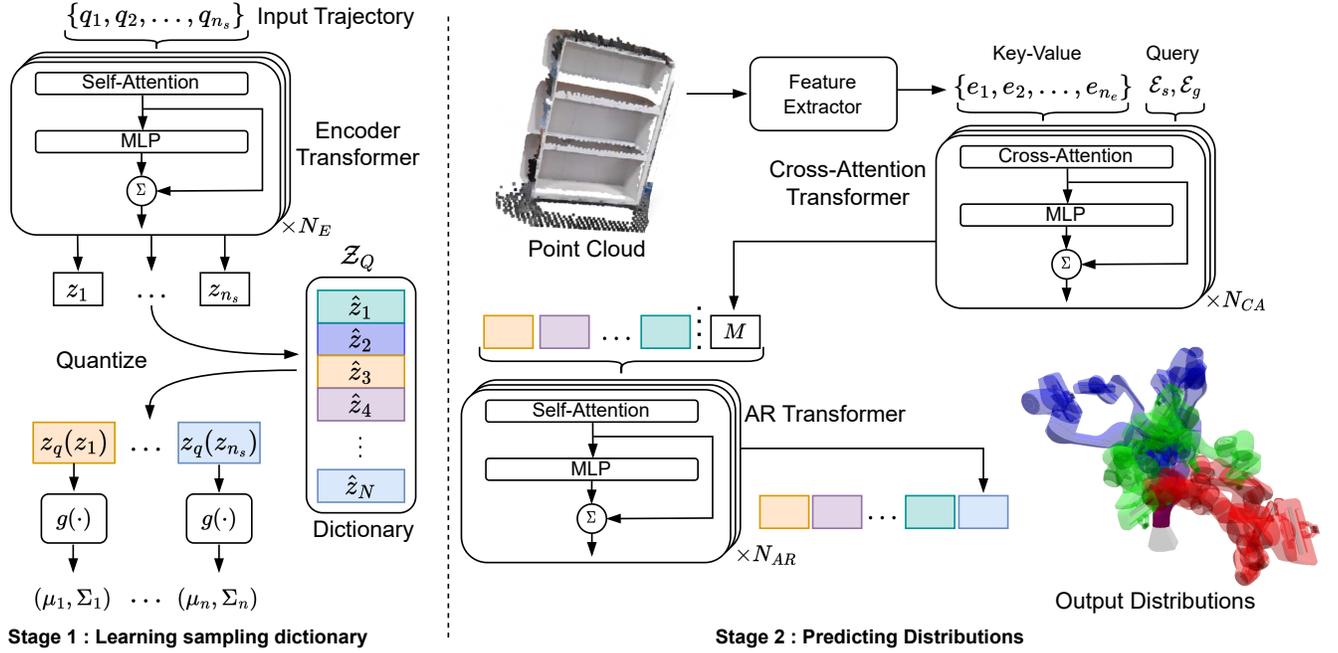}
    \caption{An outline of the model architecture of VQ-MPT. Stage 1 (Left) is a Vector Quantizer that learns a set of latent dictionary values that can be mapped to a distribution in the planning space. By encoding the planning space to discrete distributions, we can plan for high-dimensional robot systems. Stage 2 (Right) is the Auto-Regressive (AR) model that sequentially predicts the sampling regions for a given environment and a start and goal configuration. The cross-attention model transduces the start and goal embeddings given the environment embedding generated using a feature extractor. The output from the AR Transformer is mapped to a distribution in the planning space using the decoder model from Stage 1.}
    \label{fig:model_fw}
    \vspace{-1.5em}
\end{figure*}

\subsection{Problem Definition}
Consider the planning space defined by $\mathcal{X}\in\mathbb{R}^n$. We define a subspace $\mathcal{X}_{free}\subset\mathcal{X}$, such that all states in $\mathcal{X}_{free}$ do not collide with any obstacle in the environment and are considered valid configuration. The objective of the motion planner is to generate a sequence of states: $\mathcal{Q} = \{ q_1, q_2, \ldots, q_{n_s}\}$ for a given start state ($q_1$) and a goal region ($\mathcal{X}_{goal}$) such that $q_i\in \mathcal{X}_{free}, \forall i\in\{1, 2, \ldots, n_s\}$, the edge connecting $q_i$ and $q_{i+1}$ is also in $\mathcal{X}_{free}$, i.e., $(1-\alpha)q_i + \alpha q_{i+1} \in \mathcal{X}_{free},\forall \alpha \in [0, 1]$, and $q_{n_s}\in\mathcal{X}_{goal}$. The sequence of states is often referred to as a trajectory or path. In this work, we are interested in a novel learning-based approach to promote efficient sampling in $\mathcal{X}$ for generating a valid, optimized trajectory.

\subsection{Vector Quantized Models}
The VQ-VAE model has been shown to compress high-dimensional spaces such as images and audio without posterior collapse observed in VAE models \cite{NIPS2017_7a98af17}.
 We utilize a VQ-VAE in a similar manner to compress the robot planning space $\mathcal{X}$. The VQ model encodes input $q\in\mathbb{R}^n$ using a function $f$ to a latent space $\mathcal{Z}$, and is quantized to a set of learned vectors $\mathcal{Z}_Q = \{\hat{z}_1, \hat{z}_2, \ldots, \hat{z}_N\}$. 
The vectors in $\mathcal{Z}_Q$ are often called codes or dictionary values in literature. The function $g$ decodes the closest vector in $\mathcal{Z}_Q$ to $f(q)$ back to the input space. 
The parameters of $f$ and $g$ and the set of vectors in $\mathcal{Z}_Q$ are estimated using self-supervised learning by minimizing the following error,
\begin{equation}
    \mathcal{L} = \mathcal{L}_{recon} + \|\text{sg}[f(q)] - \hat{z}\| + \beta \|f(q)-\text{sg}[\hat{z}]\|,
    \label{eqn:vq_loss}
\end{equation}
where $\hat{z}$ is the quantized vector and $\text{sg}[\ ]$ stands for the stop gradient operator \cite{NIPS2017_7a98af17}, which has zero partial derivatives, i.e. $\nabla \text{sg}(x) = 0$, preventing the operand from being updated during training. 
$\mathcal{L}_{recon}$ is the main AE reconstruction loss (we will derive this later). The second term is used to update the latent vectors in $\mathcal{Z}_Q$ while keeping the encoder output constant, and the last term is called the commitment loss and updates the encoder function while keeping the latent vectors constant. This prevents the output of the encoder from drifting away from the current set of latent vectors. Yu et al. \cite{yu2022vectorquantized} proposed two further improvements in representing the codes to help improve the training stability, code usage, and reconstruction quality of VQ-VAE models for images.
\subsubsection{Factorized Codes} The output from the encoder function is linearly projected to a lower dimensional space. For example, if the encoder output is a 1024-d vector, it is projected to an 8-d vector. The authors in \cite{yu2022vectorquantized} show that using a lower dimensional space improves code usage and reconstruction quality.
\subsubsection{Normalized Codes} Each factorized codes, $\hat{z}_i$, are $l_2$ normalized. Hence all the dictionary values are mapped onto a hypersphere. This improves the training stability and reconstruction quality of the model.

\subsection{Transformer Models}\label{sec:tf_model}
Transformer models are transduction models that consist of self-attention \cite{DBLP:conf/iclr/LinFSYXZB17} and fully connected layers. They have been shown to efficiently model sequence data for language and image tasks \cite{NIPS2017_3f5ee243, dosovitskiy2021an}, hence an ideal encoder model. The self-attention layer is a Scaled Dot-product Attention \cite{NIPS2017_3f5ee243} that takes three matrices - query ($Q\in \mathbb{R}^{n_s\times d_q}$), value ($V\in\mathbb{R}^{n_s\times d_v}$), and key ($K\in\mathbb{R}^{n_s\times d_q}$) vectors to generate the attention output
\begin{equation}
    \text{Atten}(Q, K, V) = \text{softmax}\left(\gamma^{-1}QK^T\right)V,
\end{equation}
where $n_s$ is the sequence length, $d_q$ is the dimension of the query space, $d_v$ is the dimension of key and value space, and $\gamma=\sqrt{d_v}$ is a scaling factor. Rather than doing a single attention function, these models linearly project the query, key, and value vectors multiple times using different learned weights and is called the multi-headed attention model. This enables the model to attend to different features present in the data. The final output is a linear combination of individual attention values evaluated on each projected set. The pooled output is passed through deep residual multilayer perceptron (MLP) networks. In \cite{DBLP:conf/icml/XiongYHZZXZLWL20}, the authors introduce Prenorm-Transformer where the inputs to the attention and MLP layers are normalized as this makes training the model more stable.



%% file: 5.vq_mpt.tex
The VQ pipelines in image generation \cite{NEURIPS2019_5f8e2fa1, yu2022vectorquantized} consist of a quantization stage and a prediction stage. We adapt this pipeline for sequence generation and represent the planning space as a collection of distributions (Fig. \ref{fig:model_fw}). Below, we describe the two stages and objectives used for training.


\subsection{Stage 1: Vector Quantizer}
The first stage learns to represent the planning space using a set of distributions. It does not take any sensor data such as costmap or pointcloud. We use a VQ model similar to VQ-VAE \cite{NIPS2017_7a98af17} with a transformer network as the encoder and propose a maximum likelihood-based reconstruction loss to learn the set of distributions. The encoder network takes in a trajectory, $\mathcal{Q} = \{q_1, q_2, \ldots, q_{n_s}\}$, and outputs a set of latent vectors, $\mathcal{Z}=\{z_1, z_2, \ldots, z_{n_s}\}$. The decoder model, an MLP model, maps the quantized encoder output to a sequence of parameterized distributions, $\{P(\cdot\ ;\ \theta_1), P(\cdot\ ; \ \theta_2), \ldots, P(\cdot\ ;\ \theta_{n_s})\}$, in the planning space. We define our reconstruction loss as follows:
\begin{equation}
     \vspace{-1mm}
    \begin{aligned}
    \begin{split}
     \mathcal{L}_{recon} &= -\sum_{j=1}^{n_s} \log(P(q_j\ ;\ \theta_j)) \\ &\phantom{{}=}
        - \lambda \sum_{j=1}^{n_s} \mathbb{E}_{q\sim \mathcal{X}}[-\log(P(q ;\ \theta_j))]
        \end{split}
    \end{aligned}
     \label{eqn:recon_eqn}
     \vspace{-1mm}
\end{equation}
where $\lambda$ is a scaling constant. The first term maximizes the likelihood of observing the input trajectory, while the second term maximizes the differential entropy. The entropy term prevents the distribution from overfitting to each batch of data because a small batch size does not cover the entire planning space. In the following paragraphs, we provide further details of our models.

The encoder model transforms each state in the trajectory into an efficient representation by learning patterns in the sequence.
Each input state, $q_j$, to the encoder is linearly projected to a latent space $\mathbb{R}^d$, and fixed position embedding \cite{NIPS2017_3f5ee243} is added to the projected output. The resulting vector is passed through multiple blocks of Prenorm-Transformer described in Section \ref{sec:tf_model} to obtain the set $\mathcal{Z}$. 
Each latent vector $z_j\in\mathcal{Z}$ is quantized to a vector from the set $\mathcal{Z}_Q=\{\hat{z}_1, \hat{z}_2, \ldots, \hat{z}_N\}$ using the function $z_q(\cdot)$ defined by:
 \begin{equation}
     z_q(z) = \hat{z}_i \quad \text{where} \quad  i = \argmin_{k\in\{1, \ldots ,N\}} \|z - \hat{z}_k\|
     \label{eqn:quant}
 \end{equation}
where $\hat{z}_i$ is the quantized vector corresponding to $q_i$.
We prepend and append the transduced set with static encodings $z_s$ and $z_g$ to indicate the start and end of the sequence, respectively. Hence the robot trajectory $\mathcal{Q}$ is transduced to $\mathcal{\hat{Z}}=\{z_s, z_q(z_1), z_q(z_2), \ldots z_q(z_{n_s}), z_g\}$.

The decoder model maps each quantized vector, $z_q(z_i)$, to the parameterized distribution $P(\cdot\ ;\ \theta_i)$. We choose the output distribution as Gaussian, but any parametric distribution, such as Gaussian Mixture Models, Exponential distributions, or Uniform distributions, can be chosen. The decoder model outputs the mean and the covariance matrix of the Gaussian distribution ($\mathcal{N}(\mu, \Sigma)$); hence it is a function of the dictionary value $z_q(z_j),\forall j \in\{1, \ldots, n_s\}$, and is represented by $\mu(z_q(z_j))$ and $\Sigma(z_q(z_j))$ respectively. We will refer to these variables as $\mu_j$ and $\Sigma_j$ for simplicity. 

To ensure that the covariance matrix always remains positive definite during training,  we decompose $\Sigma_j$ using Cholesky decomposition as in previous works \cite{7353798, 8461047}:
\begin{equation}
    \Sigma_j = L_jD_jL_j^T
\end{equation}
where $L_j$ is a lower triangle matrix with ones along the diagonal, and $D_j$ is a diagonal matrix with positive values. The output from the penultimate MLP layer is passed through separate linear layers to obtain $\mu_j$  and $L_j$, while for $D_j$, it is passed through a linear and soft-plus layer \cite{NIPS2000_44968aec} to ensure values are positive. Using the soft-plus layer improves the stability of training the model.

\subsection{Stage 2: Auto-Regressive (AR) Prediction}
The second stage generates sampling regions by predicting indexes from the dictionary set $\mathcal{Z}_Q$ for a given planning problem and sensor data.
It comprises two models - a cross-attention model to embed start and goal pairs and the environment embedding into latent vectors ($M$), and a Transformer-based Auto-Regressive (AR) model to predict the dictionaries indexes, $\mathcal{H} = \{h_1, h_2, \ldots h_{n_h}\}$. Both models are trained end-to-end by reducing the cross entropy loss using trajectories from an RRT$^*$ planner:
\begin{equation}
    \mathcal{L}_{CE} = \mathbb{E}[-\sum_{j=1}^{n_h}\sum_{i=1}^{N+1} \delta_i(h_j) \log(\pi(h_j=i| \hat{z}_{h_1}, \cdots, \hat{z}_{h_{j-1}}, M))]
    \label{eqn:ce_loss}
\end{equation}
where $\delta_i(\cdot)$ is the Kronecker delta function, $\pi(\cdot)$ is the output of the AR model, and $\hat{z}_{h_i}$ corresponds to the latent dictionary vector associated with the ground truth index $h_i$, and the expectation is over multiple trajectories. We provide more details of the models in the following section.

The environment representation (i.e., costmap or point cloud data) is passed through a feature extractor to construct the environment encodings $\mathcal{E}=\{e_1, e_2, \ldots, e_{n_e}\}$ where $e_i \in\mathbb{R}^d$. The feature extractor reduces the dimensionality of the environment representation and captures local environment structures as latent variables using convolutional layers for costmaps and set-abstraction layers for point clouds. The start and goal states ($q_s$ and $q_g$) are projected to the start and goal embedding ($\mathcal{E}_s\in\mathbb{R}^d$ and $\mathcal{E}_g\in\mathbb{R}^d$) using a MLP network.
The cross-attention model is a Prenorm-Transformer model that uses the environment embedding, $\mathcal{E}$, and the start and goal embedding, $\{\mathcal{E}_s, \mathcal{E}_g\}$ to generate latent vectors $M$. The cross-attention model learns a feature embedding that fuses the given start and goal pair with the given planning environment. It uses the vector in $\mathcal{E}$ as key-value pairs, and $\mathcal{E}_s$ and $\mathcal{E}_g$ as query vectors to generate $M$.

We use an AR Transformer model, $\pi(\cdot)$, to predict the dictionary indexes $\mathcal{H}$. A Transformer-based AR model was chosen because of their ability to make long-horizon connections.
For each index $h_j$, the model outputs a probability distribution over $\mathcal{Z}_Q\cup \{z_g\}$ given dictionary values of previous predictions $\{\hat{z}_{h_1}, \hat{z}_{h_2}, \ldots, \hat{z}_{h_{j-1}}\}$ and the planning context $M$:
\vspace{-.5em}
\begin{equation}
\vspace{-.5em}
    \pi(h_j=i|\hat{z}_{h_1}, \ldots, \hat{z}_{h_{j-1}}, M) = p_i \quad \text{where} \ \sum_{i=1}^{N+1} p_i = 1
    \label{eqn:ar_prob}
\end{equation}
Using the learned decoder from Stage 1, we can convert each of the predicted dictionary values, $\hat{z}_{h_j}$, into a Gaussian distribution ($\mathcal{N}(\mu_{h_j}, \Sigma_{h_j})$) in the planning space.

\subsection{Generating Distributions for Sampling}
With stage 1, we have efficiently split the planning space into a discrete set of distributions represented using a set of latent vectors, and with stage 2, we have provided a means to select a subset of distributions from the dictionary.
Given a new planning problem, we use the trained Stage 2 models to generate a sequence of dictionary indexes $\mathcal{H}=\{h_1, \ldots h_{n_h}\}$. Since each index can take $N$ values, we pick the sequence $\mathcal{H}$ that maximizes the following probability:
\begin{equation}
    P(h_1, \ldots, h_{n_h} | M) = \prod_{i=1}^{n_h} \pi(h_i|h_1, \ldots, h_{i-1}, M)
    \label{eqn:prob_eqn}
\end{equation}
where $h_{n_h}$ is the goal index and $\pi$ is the probability from Eqn. \ref{eqn:ar_prob}. We apply a beam-search algorithm to optimize for Eqn. \ref{eqn:prob_eqn} as done before in language model tasks \cite{DBLP:conf/naacl/DevlinCLT19}. 

\begin{figure}
    \centering
    \includegraphics[width=0.8\linewidth]{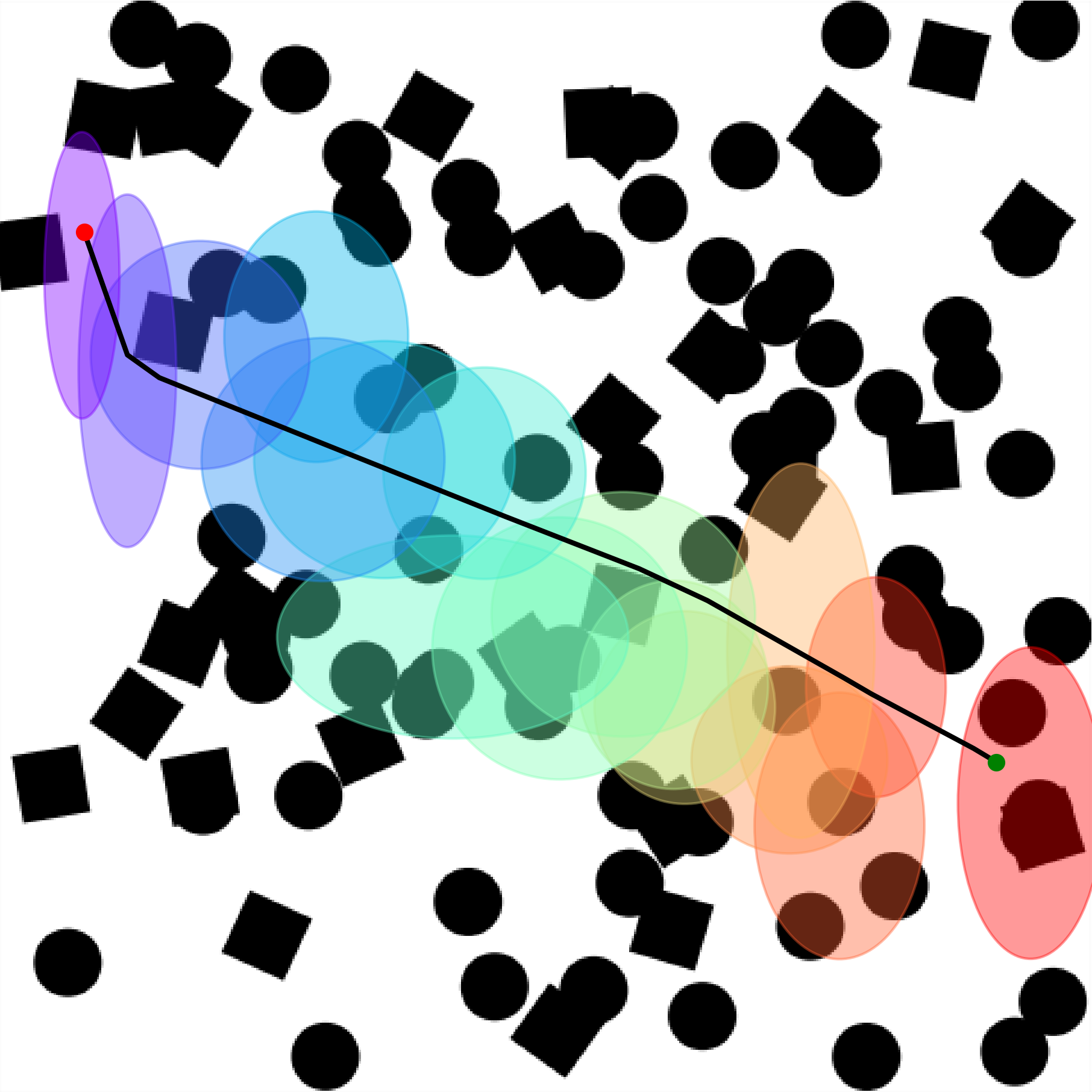}
    \caption{A trajectory (black) planned using VQ-MPT for the 2D robot and the corresponding GMM used for sampling. Each ellipse represents the distribution encoded by the dictionary values. The shaded region represents the 2 standard deviation confidence interval region. The dictionary values can encode the planning space using a finite number of vectors.}
    \label{fig:sampling_distribution}
    \vspace{-1em}
\end{figure}
The decoder model from Stage 1 is used to generate a set of distributions, $\mathcal{P}$, from the dictionary values, $\{\hat{z}_{h_1}, \hat{z}_{h_2}, \ldots, \hat{z}_{h_{n_h-1}}\}$, corresponding to the predicted indexes $\{h_1, h_2, \ldots, h_{n_h-1}\}$. We define this set as a Gaussian Mixture Model (GMM) with uniform mixing coefficients:
\begin{equation}
    \mathcal{P}(q) = \sum_{i=1}^{n_h-1} \frac{1}{n_h - 1} \mathcal{N}(\mu(\hat{z}_{h_i}), \Sigma(\hat{z}_{h_i}))
    \label{eqn:sample_dist}
\end{equation}
An example of this distribution is in Fig. \ref{fig:sampling_distribution} for a 2D robot.

\begin{table}
    \centering
    \caption{Model and environment parameters for each robot}
    \begin{tabular}{cccccc}
        \toprule
         Robot &  Environment & $d$ & Dictionary & $d_k$ & $d_v$ \\
               & Representation &   & Keys       &       &       \\
        \midrule
         2D   & Costmap       & 512 & 1024       & 512   & 256   \\
         7D   & Point Cloud   & 512 & 2048       & 512   & 256   \\
         14D  & Point Cloud   & 512 & 2048       & 512   & 256   \\
        \midrule
    \bottomrule
    \end{tabular}
    \label{tab:model_param}
    \vspace{-1.25em}
\end{table}

\begin{figure*}[t]
    \centering
    \includegraphics[width=\linewidth]{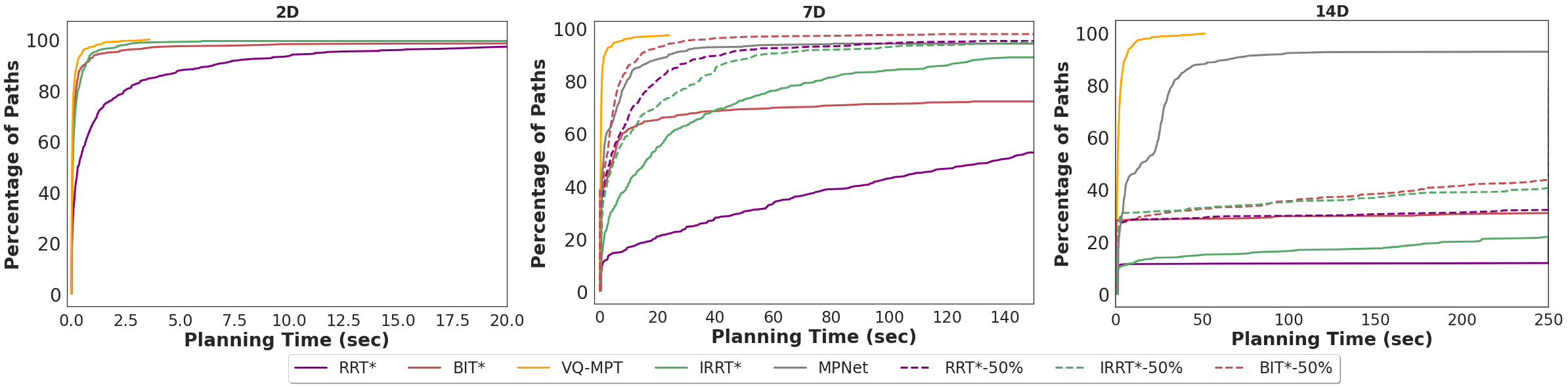}
    \caption{Plots of planning time and percentage of paths successfully planned on in-distribution environments for the 2D (Left), 7D (Center), and 14D (Right) robots. VQ-MPT can solve problems faster than other SMP planners by reducing the planning space and scales to higher dimensional problems.}
    \label{fig:percentage_complete}
\end{figure*}

\begin{table*}
    \caption{Comparing accuracy and mean planning time and vertices in In-Distribution environments }
    \centering
    \begin{tabular}{ccccccccccc}
        \toprule
        Robot & & RRT$^*$ & RRT$^*$ (50\%)& IRRT$^*$ & IRRT$^*$ (50\%) & BIT* & BIT* (50\%) & MPNet & VQ-MPT \\
        \midrule
        \multirow{2}{*}{2D}              & Accuracy   & 94.8\%  & $\cdot$ & 97.4\%  & $\cdot$ & 96.0 \% & $\cdot$ & 92.35\%   & \textbf{97.6\%} \\
                                         & Time (sec) & 1.588   & $\cdot$ & 0.244   & $\cdot$ & 0.297   & $\cdot$ & 0.296     & \textbf{0.147} \\
                                         & Vertices   &  1195   & $\cdot$ & 195     & $\cdot$ & 457     & $\cdot$ & \textbf{63}        & 306   \\
        \midrule
        \multirow{2}{*}{7D}              & Accuracy   & 52.80\%  & 95.20\% & 89.0\%  & 94.80\% & 72.20\% & 97.40\% & 94.2\%  & \textbf{97.4\%} \\
                                         & Time (sec) & 49.35    & 10.51   & 54      & 15.03   & 7.58    & 5.26  & 5.18     & \textbf{0.929} \\
                                         & Vertices   & 683      & 149     & 63      & 71      & 826     & 640   & 147      & \textbf{45}    \\
        \midrule
        \multirow{2}{*}{14D}             & Accuracy   & 11.80\%  & 32.00\% & 21.80\%  & 40.40\% & 30.80\% & 43.40\% & 92.20\%  & \textbf{99.20\%} \\
                                         & Time (sec) & 1.80     & 15.03   & 52.84    & 29.16   & 9.56    & 39.09   & 17.46    & \textbf{2.62}  \\
                                         & Vertices   & 9        & 94      & 45       & 77      & 384     & 2021    & 117      & \textbf{18}    \\
        \midrule
        \bottomrule
        \end{tabular}
        \label{tab:stats_table}
    \vspace{-1em}
\end{table*}

\begin{algorithm}[t]
\caption{VQMPTPlanner($q_{s}$, $q_{g}$, $\mathcal{P}$, $K$, $b$)}\label{alg:VQMPTPlanner}
\footnotesize
$\tau \gets \{q_s\}$\;
\For{$k\gets 0$ \KwTo $K$}{
    $q_{rand}\gets$ SAMPLE($\mathcal{P}$)\;
    $q_{near}\gets$NEAREST($q_{rand}$, $\tau$)\;
    \If{CONNECT($q_{rand}$, $q_{near}$)}{
        $\tau \gets \tau \cup \{q_{rand}\}$\;
    }
    \If{rand()$>b$}{ \label{algo: goal_check}
        $q_{gn}\gets $NEAREST($q_g$, $\tau$)\;
        \If{CONNECT($q_{gn}$, $q_g$)}{
            $\tau \gets \tau \cup \{q_{g}\}$\;
            break\;
        }
    }
    SIMPLIFY($\tau$)\;
    \Return{$\tau$}
}
\end{algorithm}
\subsection{Planning}
To generate a trajectory, any SMP can be used to generate the trajectory by sampling from the distribution given in Eqn. \ref{eqn:sample_dist}.
We use Algorithm \ref{alg:VQMPTPlanner}, to generate a path using samples from the distribution in Eqn. \ref{eqn:sample_dist}. 
The \textit{VQMPTPlanner} function takes the start and goal state ($q_s$ and $q_g$), the number of samples to generate ($K$), and a threshold value ($b$) to sample the goal state and returns a valid trajectory. This function is a modified RRT algorithm, where instead of  \textit{CONNECT} extending the current node by a small range, it checks if a valid path exists between the current and sampled node.


%% file: 6.Method.tex


We evaluated our framework on three environments - a 2D point robot, a 7D Franka Panda Arm, and a 14D Bimanual Setup.  Our experiments compare the use of VQ-MPT coupled with RRT (Algorithm \ref{alg:VQMPTPlanner}) with traditional and learning-based planners on a diverse set of planning problems. All planners were implemented using the Open Motion Planning Library (OMPL) \cite{sucan2012the-open-motion-planning-library}.

\subsection{Setup}
We trained a separate VQ-MPT model for each robot system and chose feature extractors based on environment representations. 
For costmaps, we used the Fully Convolutional Network (FCN) as in \cite{mpt}, while for point cloud data, we used two layers of set-abstraction proposed in PointNet++ \cite{NIPS2017_d8bf84be}. We chose these architectures because they are agnostic to the environment size and can generate latent embeddings for larger-sized costmaps or point clouds.
The same transformer model architecture was used for the Stage 1 encoder, the cross-attention network, and the AR model. Each transformer model consisted of 3 attention layers with 3 attention heads each. Table \ref{tab:model_param} details the latent vector dimensions and the dictionary size used for each robot. A larger key size was used for the 7D and 14D robots because of the larger planning space. We observed that increasing the dictionary size further did not reduce the reconstruction loss.

All models were trained using data collected from simulation. We collected two sets of trajectories. 
\subsubsection{Trajectories without obstacles}
This set consisted of trajectories in an environment without obstacles and was used to train Stage 1 of the model. These trajectories were free from any form of self-collision and covered the whole planning space of the planner. For each robot, we collected 2000 trajectories of this type. 
\subsubsection{Trajectories with obstacles}\label{sec:obstacle_env}
This set consisted of valid trajectories collected from environments where obstacles were placed randomly in the scene. It was used to train Stage 2 of the model. For each robot, we collected 10 trajectories for 2000 randomly generated environments.

We trained Stages 1 and 2 using the Adam optimizer \cite{DBLP:journals/corr/KingmaB14} with $\beta_1=0.9$, $\beta=0.98$ and $\epsilon=10^{-9}$ and a scheduled learning rate from \cite{NIPS2017_3f5ee243}. 

\begin{figure*}[t]
    \centering
    \includegraphics[width=\linewidth]{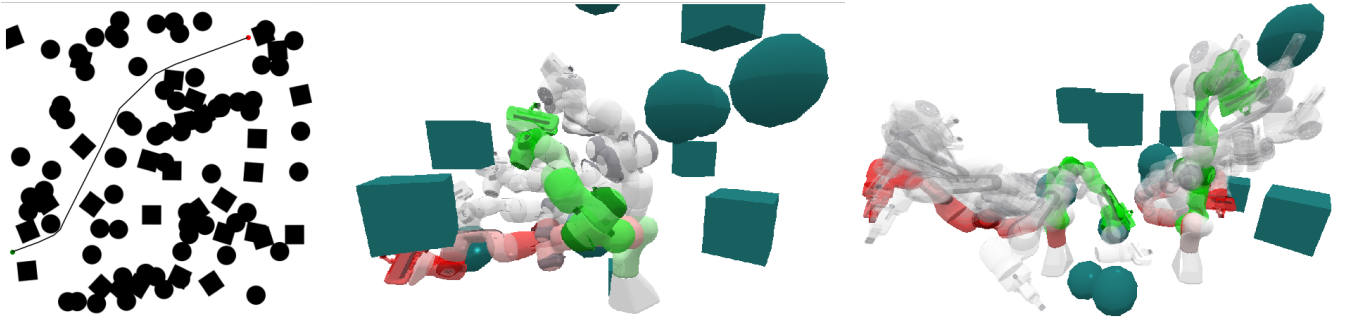}
    \caption{Sample paths planned by the VQ-MPT planner for different robot systems (Left) 2D robot, (Center) 7D robot, and (Right) 14D robot on in-distribution environments. The red and green color represents the start and goal states of the robot, respectively. Given an environment with crowded obstacles, VQ-MPT can sample efficiently from learned distributions to find a trajectory.}
    \label{fig:example_paths}
\end{figure*}


\begin{figure*}
    \centering
    \includegraphics[width=\linewidth]{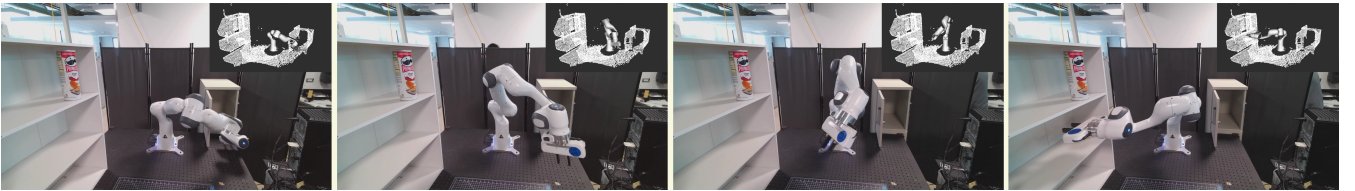}
    \caption{Snapshots of a trajectory planning using VQ-MPT for physical panda robot arm for a given start and goal pose on a shelf environment. 
    On the top-right of each image, we show the point cloud data captured using Azure Kinect cameras. 
    We used markerless camera-to-robot pose estimation to localize the captured point cloud in the robot's reference frame.
    VQ-MPT can generalize to real-world sensor data without additional training or fine-tuning.}
    \label{fig:real_shelf_plan}
    \vspace{-1.5em}
\end{figure*}

\begin{table*}[ht]
    \caption{Comparing accuracy and mean planning time and vertices in Out-of-Distribution Environments}
    \centering
    \begin{tabular}{ccccccccccc}
        \toprule
        Robot & & RRT$^*$ & RRT$^*$ (50\%) & IRRT$^*$ & IRRT$^*$ (50\%) & BIT* & BIT* (50\%) & RRT & MPNet & VQ-MPT \\
        \midrule
        \multirow{2}{*}{7D}              & Accuracy   & 8.60\% & 66.60\% & 44.60\% & 59.20\% & 37.80\% & 88.60\% & 84.20\% & 53.20\% & \textbf{92.20\%} \\
                                         & Time (sec) & 107.75 & 22.75   & 55.12   & 23.94   & 75.32   & 11.86   & 8.88    & 10.14   &  \textbf{3.24}\\
                                         & Vertices   & 1338   & 279     & 215     & \textbf{72}      & 5147    & 896     & 477     & 310     & 306 \\
        \midrule
        \multirow{2}{*}{14D}             & Accuracy   & 6.00\% & 18.60\% & 10.60\% & 17.80\% & 12.20\% & 30.00\% & 75.00\% & 80.40\% & \textbf{98.60\%} \\
                                         & Time (sec) & 4.92   & 7.61    & 20.72   & 10.57   & 30.07   & 40.58   & 19.75   & 23.91   & \textbf{6.21}\\
                                         & Vertices   & 39     & 67      & 20      & 34      & 1673    & 2889    & 179     & 104     & \textbf{70}\\
        \midrule
        \multirow{3}{*}{7D (Real)}       & Accuracy   &$\cdot$ &$\cdot$  & 100\%   &$\cdot$  & 100\%   &$\cdot$  & 100\%   & 30\%    & 100\% \\
                                         & Time (sec) &$\cdot$ &$\cdot$  & 30.68   &$\cdot$  & 26.42   &$\cdot$  & 1.69    & 2.23    & \textbf{1.17} \\
                                         & Vertices   &$\cdot$ &$\cdot$  & 607     &$\cdot$  & 2852    &$\cdot$  & \textbf{21} & 7   & 34 \\
        \midrule
    \bottomrule
    \end{tabular}
    \label{tab:untrain_stats_table}
\end{table*}

\begin{figure*}
    \centering
    \includegraphics[width=0.8\linewidth]{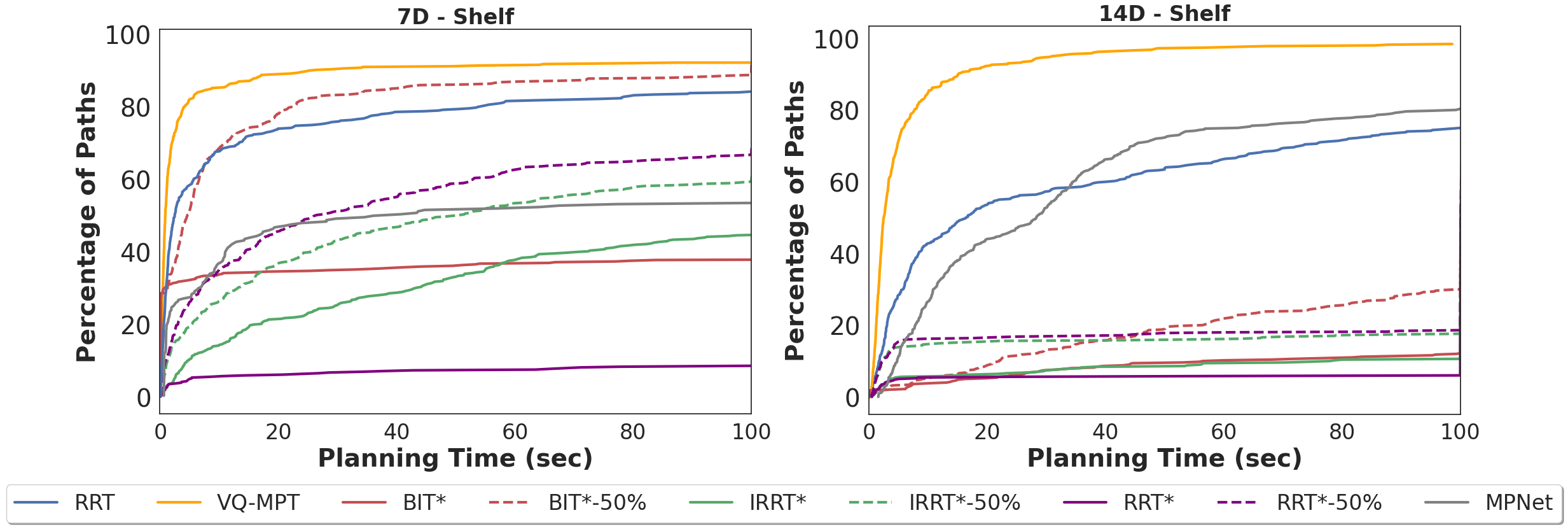}
    \caption{Plots of planning time and percentage of paths successfully planned for the 7D (Left) and 14D (Right) robots on environments different from ones used for training. VQ-MPT can reduce the planning space in unseen environments, enabling efficient planning in challenging environments.}
    \label{fig:percentage_solved_shelf}
    \vspace{-1em}
\end{figure*}
\subsection{Results - Unseen In-Distribution Environments} \label{sec:results}
We compared our framework against traditional and learning-based SMP algorithms for each robot system on a trajectory from 500 different environments. 
To quantify planning performance, we measured three metrics: planning time - the time it takes for the planner to generate a valid trajectory; vertices - the number of collision-free vertices required to find the trajectory and accuracy - the percentage of planning problems solved before a given cutoff time. We chose to measure vertices because checking the validity of a vertex imposes a significant cost on most SMPs \cite{9023003}. Since optimal planners do not have termination conditions, for fair comparisons, we stopped planning when the constructed trajectory, $\{q_1, q_2, \ldots, q_n\}$, satisfied the following condition:
\begin{equation}
    \sum_{i=0}^{n-1} \|q_{i+1}-q_i \|_2 \leq (1+\epsilon) \sum_{j=0}^{m-1}\|q_{j+1}^* - q_j^* \|_2 
    \label{eqn:plan_constraint}
\end{equation}
where $\mathcal{Q}^* = \{q_1^*, \ldots, q_n^*\}$ is the path planned by VQ-MPT and $\epsilon \geq 0$ is a user-defined threshold.
If VQ-MPT could not generate a path for the trajectory, we used a path from RRT$^*$ running for 300 seconds (s) to generate $\mathcal{Q}^*$. For optimal planners like RRT$^*$, IRRT$^*$, and BIT$^*$, we used $\epsilon=0.1$ and $\epsilon=0.5$. In our tables, planners that used $\epsilon=0.5$ are reported by `X (50\%)', where X is the planner. The planning time reported for VQ-MPT also includes the time taken for model inference. All results are summarized in Table \ref{tab:stats_table} and the percentage of planning problems solved vs planning time is shown in Figure \ref{fig:percentage_complete}.

We first tested our framework on a simple 2D robot. An example of the path planned by the VQ-MPT framework is shown in Fig. \ref{fig:example_paths} (Left).  The cutoff time set was 20 seconds. VQ-MPT showed efficient sampling of points in the planning space and found trajectories faster than traditional planners.


VQ-MPT can also use 3D environment representations such as point clouds to generate sampling regions. We evaluated the framework on a 7D panda robot arm with a point cloud environment representation. The dictionary encodings can capture diverse sets of valid configurations in 7D space (Fig. \ref{fig:model_fw}). An example of the trajectory planned by the VQ-MPT framework is shown in Fig. \ref{fig:example_paths} (Center). The cutoff time set was 100 s. VQ-MPT planner generates a trajectory nearly 5$\times$ faster with fewer vertices than the next best accurate planner. 
MPNet performs poorly compared to VQ-MPT. The rigid feature encoding of MPNet potentially prevents it from generalizing to larger point cloud data environments. VQ-MPT, in contrast, learns to identify suitable regions to sample in the joint space using point cloud data of different sizes.

We also tested the framework in a bi-manual panda arm setup with 14D. An example of a VQ-MPT trajectory is shown in Fig. \ref{fig:example_paths} (Right). Stage 1 captures the planning space with the same 2048 dictionary values used in the 7D panda experiment. The cutoff time was 250 s. 
While BIT* performed relatively well compared to traditional planners for the 2D and 7D problems, performance and accuracy decreased due to the high-dimensional planning space. Since Stage 1 of the VQ-MPT framework encodes self-collision-free regions, it's easier for the planner to generate feasible trajectories in Stage 2, resulting in faster trajectory generation with fewer vertices.

\subsection{Results - Out-of-Distribution Environments}
Our next set of experiments evaluated VQ-MPT's performance for the 7D and 14D robots in environments very different from the training environments. We test our framework on different planning scenes resembling real-world scenarios (Fig. \ref{fig:sample_fig}). We test the model for each robot on 500 and 10 start and goal locations for simulation and real-world environments, respectively. The cutoff time for each planner was set at 100 s.
The results of the experiments are summarized in Table \ref{tab:untrain_stats_table}, and the plot of the percentage of paths solved across planning time is given in Fig. \ref{fig:percentage_solved_shelf}. 
Higher dimensional 7D and 14D spaces are challenging. The environment is even more challenging because of the goal location inside the shelf since it reduces the number of feasible trajectories in the same way a narrow passage eliminates feasible trajectories in mobile robots \cite{88137}. Even non-optimal planners like RRT solve only 75-91\% of trajectories. 
Existing optimal SMP planners cannot achieve the same accuracy as VQ-MPT even after relaxing path length constraints. 

To evaluate the performance of VQ-MPT on physical sensor data, we tested a trained model in a real-world environment (Fig. \ref{fig:real_shelf_plan}). The environment was represented using point cloud data from Azure Kinect sensors, and collision checking was done using the octomap collision checker from Moveit \footnote{https://moveit.ros.org/}. Camera to robot base transform was estimated using markerless pose estimation technique \cite{lu2023markerless}. Our results show that the model can plan trajectories faster than RRT with the same accuracy. We observed that VQ-MPT trajectories are also shorter than RRT trajectories, which can be clearly seen in some of the attached videos. This experiment shows that VQ-MPT models can also generalize well to physical sensor data without further training or fine-tuning. Such generalization will benefit the larger robotics community since other researchers can use trained models in diverse settings without collecting new data or fine-tuning the model.

%% file: 8.Conclusion.tex


VQ-MPT can plan near-optimal paths in a fraction of the time required by traditional planners, scales to higher dimension planning space, and achieves better generalizability than previous learning-based planners. Our approach will be beneficial for planning multi-arm robot systems like the ABB Yumi and Intuitive's da Vinci\textregistered Surgical System. It is also helpful for applications where generating nodes and edges for SMPs is computationally expensive, such as for constrained motion planning \cite{9551655}. Future works will extend VQ-MPT to these applications.

%% file: root.bbl
\begin{thebibliography}{10}
\providecommand{\url}[1]{#1}
\csname url@samestyle\endcsname
\providecommand{\newblock}{\relax}
\providecommand{\bibinfo}[2]{#2}
\providecommand{\BIBentrySTDinterwordspacing}{\spaceskip=0pt\relax}
\providecommand{\BIBentryALTinterwordstretchfactor}{4}
\providecommand{\BIBentryALTinterwordspacing}{\spaceskip=\fontdimen2\font plus
\BIBentryALTinterwordstretchfactor\fontdimen3\font minus
  \fontdimen4\font\relax}
\providecommand{\BIBforeignlanguage}[2]{{%
\expandafter\ifx\csname l@#1\endcsname\relax
\typeout{** WARNING: IEEEtran.bst: No hyphenation pattern has been}%
\typeout{** loaded for the language `#1'. Using the pattern for}%
\typeout{** the default language instead.}%
\else
\language=\csname l@#1\endcsname
\fi
#2}}
\providecommand{\BIBdecl}{\relax}
\BIBdecl

\bibitem{doi:10.1177/02783640122067453}
S.~M. LaValle and J.~James J.~Kuffner, ``Randomized kinodynamic planning,''
  \emph{The International Journal of Robotics Research}, 2001.

\bibitem{508439}
L.~Kavraki, P.~Svestka, J.-C. Latombe, and M.~Overmars, ``Probabilistic
  roadmaps for path planning in high-dimensional configuration spaces,''
  \emph{IEEE Trans. on Robotics and Auto.}, 1996.

\bibitem{1242285}
D.~Hsu, T.~Jiang, J.~Reif, and Z.~Sun, ``The bridge test for sampling narrow
  passages with probabilistic roadmap planners,'' in \emph{IEEE Int. Conf. on
  Robotics and Auto.}, 2003.

\bibitem{9561673}
Z.-Y. Chiu, F.~Richter, E.~K. Funk, R.~K. Orosco, and M.~C. Yip, ``Bimanual
  regrasping for suture needles using reinforcement learning for rapid motion
  planning,'' in \emph{IEEE Int. Conf. on Robotics and Auto.}, 2021.

\bibitem{1570348}
R.~Alterovitz, K.~Goldberg, and A.~Okamura, ``Planning for steerable bevel-tip
  needle insertion through 2d soft tissue with obstacles,'' in
  \emph{Proceedings of the IEEE Int. Conf. on Robotics and Auto.}, 2005.

\bibitem{6942976}
J.~D. Gammell, S.~S. Srinivasa, and T.~D. Barfoot, ``Informed {RRT}*: Optimal
  sampling-based path planning focused via direct sampling of an admissible
  ellipsoidal heuristic,'' in \emph{Int. Conf. on Intelligent Robots and
  Systems}, 2014.

\bibitem{gammell2015batch}
------, ``{B}atch informed trees ({BIT*}): Sampling-based optimal planning via
  the heuristically guided search of implicit random geometric graphs,'' in
  \emph{2015 IEEE Int. Conf. Robot. Autom.}, 2015.

\bibitem{qureshi2016potential}
A.~H. Qureshi and Y.~Ayaz, ``Potential functions based sampling heuristic for
  optimal path planning,'' \emph{Autonomous Robots}, 2016.

\bibitem{tahir2018potentially}
Z.~Tahir, A.~H. Qureshi, Y.~Ayaz, and R.~Nawaz, ``Potentially guided
  bidirectionalized rrt* for fast optimal path planning in cluttered
  environments,'' \emph{Robotics and Autonomous Systems}, 2018.

\bibitem{8412538}
P.~Lehner and A.~Albu-Schäffer, ``The repetition roadmap for repetitive
  constrained motion planning,'' \emph{IEEE Robot. and Autom. Letters}, 2018.

\bibitem{9561104}
C.~Chamzas, Z.~Kingston, C.~Quintero-Peña, A.~Shrivastava, and L.~E. Kavraki,
  ``Learning sampling distributions using local 3d workspace decompositions for
  motion planning in high dimensions,'' in \emph{IEEE Int. Conf. on Robot. and
  Autom.}, 2021.

\bibitem{8653875}
B.~Ichter and M.~Pavone, ``Robot motion planning in learned latent spaces,''
  \emph{IEEE Robotics and Auto. Letters}, 2019.

\bibitem{kumar2019lego}
R.~Kumar, A.~Mandalika, S.~Choudhury, and S.~Srinivasa, ``Lego: Leveraging
  experience in roadmap generation for sampling-based planning,'' in \emph{Int.
  Conf. on Intelligent Robots and Systems}, 2019.

\bibitem{qureshi2019motion}
A.~H. Qureshi, Y.~Miao, A.~Simeonov, and M.~C. Yip, ``Motion planning networks:
  Bridging the gap between learning-based and classical motion planners,''
  \emph{IEEE Trans. on Robotics}, 2020.

\bibitem{mpt}
J.~J. Johnson, U.~S. Kalra, A.~Bhatia, L.~Li, A.~H. Qureshi, and M.~C. Yip,
  ``Motion planning transformers: A motion planning framework for mobile
  robots,'' 2021.

\bibitem{DBLP:conf/iclr/ChenDLYLS20}
B.~Chen, B.~Dai, Q.~Lin, G.~Ye, H.~Liu, and L.~Song, ``Learning to plan in high
  dimensions via neural exploration-exploitation trees,'' in \emph{Int. Conf.
  on Learning Representations, {ICLR}}, 2020.

\bibitem{yu2021reducing}
C.~Yu and S.~Gao, ``Reducing collision checking for sampling-based motion
  planning using graph neural networks,'' in \emph{Advances in Neural
  Information Processing Systems}, 2021.

\bibitem{NIPS2017_3f5ee243}
A.~Vaswani, N.~Shazeer, N.~Parmar, J.~Uszkoreit, L.~Jones, A.~N. Gomez, L.~u.
  Kaiser, and I.~Polosukhin, ``Attention is all you need,'' in \emph{Advances
  in Neural Information Processing Systems}, 2017.

\bibitem{DBLP:conf/naacl/DevlinCLT19}
J.~Devlin, M.~Chang, K.~Lee, and K.~Toutanova, ``{BERT:} pre-training of deep
  bidirectional transformers for language understanding,'' in \emph{Proceedings
  of the Conference of the North American Chapter of the Association for
  Computational Linguistics: Human Language Technologies}, 2019.

\bibitem{NEURIPS2020_1457c0d6_GPT3}
T.~Brown, B.~Mann, N.~Ryder, M.~Subbiah, J.~D. Kaplan, P.~Dhariwal,
  A.~Neelakantan, P.~Shyam, G.~Sastry, A.~Askell, S.~Agarwal, A.~Herbert-Voss,
  G.~Krueger, T.~Henighan, R.~Child, A.~Ramesh, D.~Ziegler, J.~Wu, C.~Winter,
  C.~Hesse, M.~Chen, E.~Sigler, M.~Litwin, S.~Gray, B.~Chess, J.~Clark,
  C.~Berner, S.~McCandlish, A.~Radford, I.~Sutskever, and D.~Amodei, ``Language
  models are few-shot learners,'' in \emph{Advances in Neural Information
  Processing Systems}, 2020.

\bibitem{chen2021decision_TransformerRL}
L.~Chen, K.~Lu, A.~Rajeswaran, K.~Lee, A.~Grover, M.~Laskin, P.~Abbeel,
  A.~Srinivas, and I.~Mordatch, ``Decision transformer: Reinforcement learning
  via sequence modeling,'' in \emph{Advances in Neural Information Processing
  Systems}, 2021.

\bibitem{janner2021sequence}
M.~Janner, Q.~Li, and S.~Levine, ``Offline reinforcement learning as one big
  sequence modeling problem,'' in \emph{Advances in Neural Information
  Processing Systems}, 2021.

\bibitem{yang2022learning}
R.~Yang, M.~Zhang, N.~Hansen, H.~Xu, and X.~Wang, ``Learning vision-guided
  quadrupedal locomotion end-to-end with cross-modal transformers,'' in
  \emph{Int. Conf. on Learning Representations}, 2022.

\bibitem{chaplot2020differentiable}
D.~S. Chaplot, D.~Pathak, and J.~Malik, ``Differentiable spatial planning using
  transformers,'' in \emph{ICML}, 2021.

\bibitem{NIPS2017_7a98af17}
A.~van~den Oord, O.~Vinyals, and k.~kavukcuoglu, ``Neural discrete
  representation learning,'' in \emph{Advances in Neural Information Processing
  Systems}, 2017.

\bibitem{yu2022vectorquantized}
J.~Yu, X.~Li, J.~Y. Koh, H.~Zhang, R.~Pang, J.~Qin, A.~Ku, Y.~Xu, J.~Baldridge,
  and Y.~Wu, ``Vector-quantized image modeling with improved {VQGAN},'' in
  \emph{Int. Conf. on Learning Representations}, 2022.

\bibitem{DBLP:conf/iclr/LinFSYXZB17}
Z.~Lin, M.~Feng, C.~N. dos Santos, M.~Yu, B.~Xiang, B.~Zhou, and Y.~Bengio, ``A
  structured self-attentive sentence embedding,'' in \emph{Int. Conf. on
  Learning Representations}, 2017.

\bibitem{dosovitskiy2021an}
A.~Dosovitskiy, L.~Beyer, A.~Kolesnikov, D.~Weissenborn, X.~Zhai,
  T.~Unterthiner, M.~Dehghani, M.~Minderer, G.~Heigold, S.~Gelly, J.~Uszkoreit,
  and N.~Houlsby, ``An image is worth 16x16 words: Transformers for image
  recognition at scale,'' in \emph{Int. Conf. on Learning Representations},
  2021.

\bibitem{DBLP:conf/icml/XiongYHZZXZLWL20}
R.~Xiong, Y.~Yang, D.~He, K.~Zheng, S.~Zheng, C.~Xing, H.~Zhang, Y.~Lan,
  L.~Wang, and T.~Liu, ``On layer normalization in the transformer
  architecture,'' in \emph{Int. Conf. on Machine Learning}, 2020.

\bibitem{NEURIPS2019_5f8e2fa1}
A.~Razavi, A.~van~den Oord, and O.~Vinyals, ``Generating diverse high-fidelity
  images with vq-vae-2,'' in \emph{Advances in Neural Information Processing
  Systems}.\hskip 1em plus 0.5em minus 0.4em\relax Curran Associates, Inc.,
  2019.

\bibitem{7353798}
H.~Hu and G.~Kantor, ``Parametric covariance prediction for heteroscedastic
  noise,'' in \emph{Int. Conf. on Intelligent Robots and Systems (IROS)}, 2015.

\bibitem{8461047}
K.~Liu, K.~Ok, W.~Vega-Brown, and N.~Roy, ``Deep inference for covariance
  estimation: Learning gaussian noise models for state estimation,'' in
  \emph{2018 IEEE Int. Conf. on Robotics and Auto. (ICRA)}, 2018, pp.
  1436--1443.

\bibitem{NIPS2000_44968aec}
C.~Dugas, Y.~Bengio, F.~B\'{e}lisle, C.~Nadeau, and R.~Garcia, ``Incorporating
  second-order functional knowledge for better option pricing,'' in
  \emph{Advances in Neural Information Processing Systems}, 2000.

\bibitem{sucan2012the-open-motion-planning-library}
I.~A. {\c{S}}ucan, M.~Moll, and L.~E. Kavraki, ``The {O}pen {M}otion {P}lanning
  {L}ibrary,'' \emph{{IEEE} Robotics \& Auto. Magazine}, 2012.

\bibitem{NIPS2017_d8bf84be}
C.~R. Qi, L.~Yi, H.~Su, and L.~J. Guibas, ``Pointnet++: Deep hierarchical
  feature learning on point sets in a metric space,'' in \emph{Advances in
  Neural Information Processing Systems}, 2017.

\bibitem{DBLP:journals/corr/KingmaB14}
D.~P. Kingma and J.~Ba, ``Adam: {A} method for stochastic optimization,'' in
  \emph{Int. Conf. on Learning Representations}, 2015.

\bibitem{9023003}
N.~Das and M.~Yip, ``Learning-based proxy collision detection for robot motion
  planning applications,'' \emph{IEEE Trans. on Robotics}, 2020.

\bibitem{88137}
J.~Borenstein and Y.~Koren, ``The vector field histogram-fast obstacle
  avoidance for mobile robots,'' \emph{IEEE Trans. on Robotics and Auto.},
  1991.

\bibitem{lu2023markerless}
J.~Lu, F.~Richter, and M.~C. Yip, ``Markerless camera-to-robot pose estimation
  via self-supervised sim-to-real transfer,'' 2023.

\bibitem{9551655}
J.~J. Johnson and M.~C. Yip, ``Chance-constrained motion planning using modeled
  distance- to-collision functions,'' in \emph{Int. Conf. on Auto. Science and
  Engineering (CASE)}, 2021.

\end{thebibliography}
